\documentclass[twocolumn]{svjour3}          
\usepackage{graphicx}
\usepackage{mathptmx}      
\usepackage{amsmath}
\usepackage{bm}
\usepackage{amssymb}
\usepackage{xspace}
\usepackage{natbib}
\usepackage{multirow}
\usepackage{grffile}		
\usepackage{epstopdf}		

\makeatletter
\DeclareRobustCommand\onedot{\futurelet\@let@token\@onedot}
\newcommand{\@onedot}{\ifx\@let@token.\else.\null\fi\xspace}

\makeatother

\begin{document}

\title{Prediction of Manipulation  Actions 
}


\author{Cornelia Ferm\"uller \and Fang Wang \and Yezhou Yang \and
         Konstantinos Zampogiannis \and Yi Zhang \and   Francisco Barranco \and Michael Pfeiffer
}


\institute{
C. Ferm\"{u}ller, Y. Yang, K. Zamgogiannis, Y. Zhang  \at University of Maryland, College Park, MD 20742, USA\\ \email{fer@cfar.umd.edu}\\  
              F. Wang  \at
              College of Engineering and Computer Science (CECS), Australian National University \\      
           F. Barranco \at
              University of Granada\\
          M. Pfeiffer \at
              Institute of Neuroinformatics,
University of Zurich and ETH Z\"urich
}


\maketitle
\begin{abstract}
Looking at a person's hands one often can tell what the person is going to do next, how his/her hands are moving  and where they will be, because an actor's intentions shape his/her movement kinematics during action execution. Similarly, active systems with real-time constraints must not simply rely on passive video-segment classification, but they have to continuously update their estimates and predict future actions.  In this paper, we study the prediction of dexterous actions. We recorded from subjects performing different manipulation actions on the same object, such as  ``squeezing'', ``flipping'', ``washing'', ``wiping'' and ``scratching'' with a sponge. In psychophysical experiments, we  evaluated  human observers' skills in predicting actions from video sequences of different length, depicting the hand movement in the preparation and execution of actions before and after contact with the object.  We then developed a recurrent neural network based method for action prediction using as input patches around the hand. We also used the same formalism  to predict the forces on the  finger tips using  for training synchronized video and force data streams.  Evaluations on two new datasets show that our system closely matches human performance in the recognition task, and demonstrate the ability of our algorithms to predict real-time what and how  a  dexterous action is performed.

\end{abstract}
\section{Introduction}


Human action and activity  understanding  has been a topic of great interest in Computer Vision and Robotics in recent  years. Many techniques have been developed for recognizing actions and large benchmark datasets have been proposed, with most of them focusing on full-body actions \citep{Mandary2015,Takano2015,Schuldt04,li2010learning,moeslund2006survey,turaga2008machine}. Typically, computationally  approaches treat  action recognition   as a classification problem, where the input is a previously segmented video, and the output a set of candidate action labels. 

\begin{figure}[t]
  \centering
 \includegraphics[width=\columnwidth]{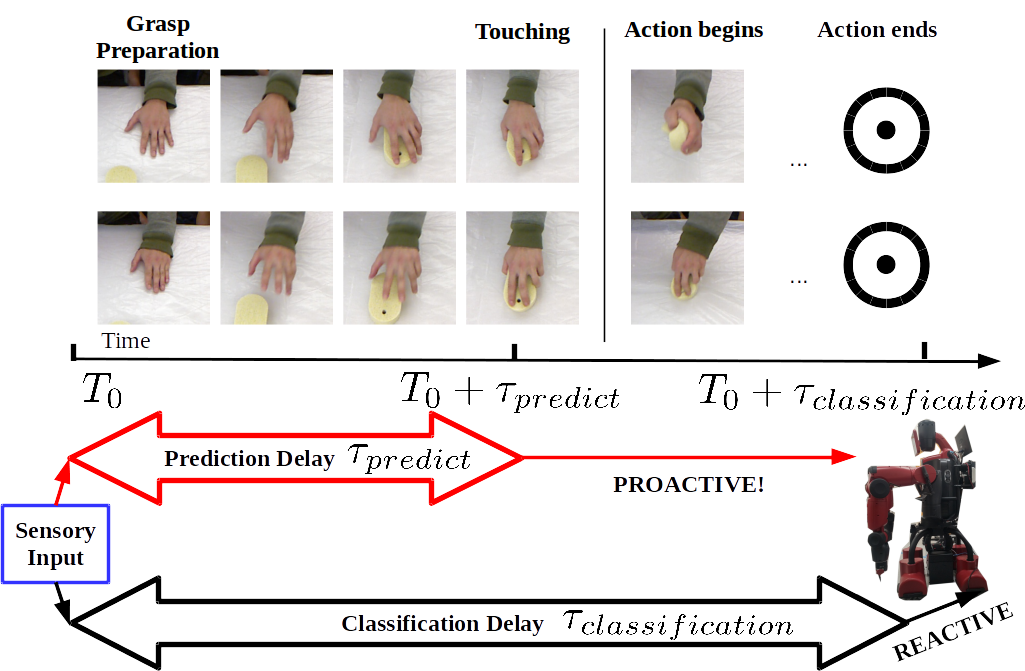}
  \caption{Two examples demonstrate that early movements are strong indicators of the intended manipulation actions. Inspired by this, our system performs action predictions from early visual cues. Compared
to the classification delay, earlier prediction of action significantly reduces
the delay in real-time interaction, which is fundamentally important
for a proactive system. (Top row: squeezing a sponge; bottom
row: wiping a table with a sponge.)}
\label{fig:action_pred}
\end{figure}

However, there is more to action understanding, as demonstrated by  biological vision. As we humans observe, we constantly perceive, and update our belief about the observed  action and about future events. We constantly recognize the ongoing action. But there is even more to it. We can understand the kinematics of the ongoing action, the limbs' future positions and velocities. We  also understand the observed actions in terms of our own motor-representations. That is, we are able to interpret others' actions in terms of dynamics and forces, and predict the effects of these forces on objects. Similarly, cognitive  robots that will assist human partners will need to  understand their intended actions at an early stage. If a robot needs to act, it cannot have a long delay in visual processing. It needs  to recognize in real-time  to plan its actions.
A fully functional perception action loop requires the robot to \emph{predict}, so it can  efficiently allocate future processes. Finally, even vision processes for multimedia tasks may benefit from being predictive.  Interpreting human activities is a very complex task and requires both, low-level vision processes and high-level cognitive processes with  knowledge about actions. \citep{Gupta2008,Kulkarni13}. Considering the challenges in state of the art visual action recognition, we argue that by integrating closely the high-level with the low-level vision processes, with the  high-level  modifying the visual processes \citep{cognitivedialogue}, a better recognition may be achieved. Prediction plays an essential component in this interaction. 
%
We can think about  the action-perception loop of our cognitive system  from the viewpoint of a control system. The sensors take measurements of the human activity. We then apply visual operations on this signal and extract (possibly using additional cognitive processes) useful information for creating the control signal in order to change the state of the cognitive system. Because the processing of the signal takes time, this creates a delay for the control \citep{Doyle2011}. It is therefore important to compute meaningful information that allows us to predict the future state of the cognitive system. In this work, we are specifically interested in manipulation actions and how  visual information of hand movements can be exploited for  predicting future action
so that the crucial delay in the control loop can be shortened (for an illustration see Fig. \ref{fig:action_pred}).

Hand movements and actions have long been studied in Computer Vision to create systems for applications such as
recognition of sign language \citep{erol2007vision}. More recent
applications include gesture recognition \citep{molchanov2015hand}, visual interfaces \citep{melax2013dynamics}, and driver analysis \citep{ohn2014hand}.  Different methods model the temporal evolution of  actions using formalisms such as  Hidden Markov models \citep{starner1998real}, Conditional Random Fields \citep{wang2006hidden} and 3d Convolutional Neural Networks \citep{molchanov2015hand}. While in principle, some of these approaches, could be used for  online prediction, they are always treated as recognition modules. In recent years a number of works have developed tools for general hand pose estimation and hand tracking, which can be  building blocks for applications involving hand movement recognition. For example, building on  work on full-body recognition \citep{shotton2013real}, \citep{keskin2013real} develops a learning-based approach using depth contrast features and Random Forest classifiers.  \cite{Oikonomidis2011} in a model-based approach use  a 27-degree of freedom model of the hand built from geometric primitives and GPU accelerated Particle Swarm Optimization. So far, these trackers and pose estimators work well on isolated hands, but methods still struggle  with hands in interaction with objects \citep{supancic2015depth}, although there are efforts underway to deal with such situations \citep{panteleris20153d}.

Inspiration for our work comes from studies in Cognitive Sciences on hand motion. The grasp and the movement kinematics are strongly related to the manipulation action \citep{jeannerod1984timing}. It has been shown that an actor's intention shapes his/her movement kinematics during movement execution, and, furthermore, observers are sensitive to this information \citep{Ansuni2015}. They can see early differences in visual kinematics and use them to discriminate between movements performed with different intentions. Kinematic studies have looked at such   physical differences in movement. For example, \cite{Ansuni2008} found that when subjects grasped  a bottle for pouring,  the middle and the ring fingers were more extended than when they grasped the bottle with the intent of displacing, throwing, or passing it. Similarly, \cite{Craje2011} found that subjects placed their thumb and index fingers in higher positions when the bottle was grasped to pour than  to lift.

It appears that the visual information in the early phases of the action is often sufficient  for observers to understand the intention of action. Starting from this intuition, we a.) conducted a study to evalute humans' performance in recognizing manipulation actions; b.) implemented a computational system using state-of the art learning algorithms. 

The  psychophysical experiment was designed to evaluate human's  performance in recognizing manipulation actions in their  early phases. These  include: \emph{1)} the grasp preparation, which is the phase when the hand moves towards the object and the fingers shape to touch the object; \emph{2)} the grasp, when the hand comes in contact with the object to hold it in a stable position;  and \emph{3)} the early actual action movement of the hand together with the object. Throughout these three phases,  observers' judgment of the action becomes more reliable and confident. The study  gives us an insight about the difficulty of the task and provides data for evaluating our computational method. 

Our computational approach processes the sensory input as a continuous signal and formulates action interpretation as a continuous updating of the prediction of intended action. This  concept is applied to two different tasks. First, from the stream of video input, we continuously predict the identity of the ongoing action. Second, using as input the video stream, we predict the forces on the fingers applied to grasped objects.
Next, we provide a  motivation for our choice of the two tasks, after which we give an overview  of  our approach.

The first task is about action prediction from video.
We humans are able to update our beliefs about the observed action, and predict it before it is completed. This capability is essential to be pro-active and react to the actions of others. Robots that interact with humans also need such capability. Predicting future actions of their counterpart allows them to allocate computational resources for their own reaction appropriately. For example, if a person is passing a cup to the robot, it has to understand what is happening well before the action is completed, so it can prepare the appropriate action to receive it. Furthermore, vision processes have to be initiated and possibly tuned with \emph{predicted} information, so the cup can be detected at the correct location, its pose estimated, and possibly other task-specific processes performed (for example, the content of the cup may need to be recognized).

The second task is about predicting the tactile signal of the intended action.
Findings of neuroscience on the mirror neuron system
\citep{Gallese98,rizzolatti2001neurophysiological}
provide evidence for a close relationship between mechanisms of action and perception in primates. Humans develop haptic perception through interaction with objects and learn to relate haptic  with visual perception. Furthermore, they develop  the capability of hallucinating the haptic stimulus when seeing hands in certain configurations  interacting with objects \citep{Bergmann2014}.
 This  capability of hallucinating force patterns from visual input is essential for a more detailed analysis of  the interaction with the physical world. It can be used to  \emph{reason about the current interaction} between the hand and the object, and to \emph{predict the action consequences} driven by the estimated force pattern. 

Furthermore, by associating vision with forces, we expect to obtain better computational action recognition modules. Intuitively, the force vectors, whose dimensions are much lower than  the visual descriptors, should provide useful compact information for classification, especially when the training data is not large. A first experiment, presented in  Section \ref{sec:predict_forces}, confirms this idea. 
 
Most important, the force patterns may be used  in robot learning. A popular paradigm in Robotics is imitation learning or learning from demonstration \citep{argall2009survey}, where the robot learns from examples provided by a demonstrator. If the  forces can be predicted from images,  then the   force profiles together with  the positional information can  be used to teach the robot  with video only. 
Many researchers are trying to teach robots actions and skills that involve forces, e.g.
wiping a kitchen table \citep{Gams2010}, pull and flip tasks \citep{Kober2015}, ironing  or opening a door \citep{Kormushev2011}. These approaches rely on haptic devices or  force and torque sensors on the robot to obtain  the force profiles for the robot to learn the  task. If we can predict the forces exerted by the human demonstrator, the demonstration could become vision only. This  would allow us to teach robots force interaction tasks much more efficiently. 


In order to solve the above two tasks, we take advantage of new developments in machine learning. Specifically, we build on the recent success of   recurrent neural networks (RNNs) in conjunction with visual features from pre-trained convolutional neural networks (CNNs) and  training  from a limited number of weakly annotated data. For the first task, we use  an RNN to recognize the ongoing action from video input. A camera records videos of humans performing a number of manipulation actions on different objects. For example, they `drink' from a cup, `pour' from it, `pound', `shake', and `move' it; or they `squeeze' a sponge, `flip' it,  `wash', `wipe', and `scratch' with it. Our system extracts patches around the hands, and feeds these patches to an RNN, which was trained offline to predict in real-time the ongoing action. For the second task, we collected videos of actions and synchronized streams of force measurements  on the hand, and we used this data to train an RNN to predict the forces, using only the segmented hand patches in  video input.

The main contributions of the paper are: \emph{1)} we present  the first computational study on the  prediction  of observed dexterous actions \emph{2)} we demonstrate an implementation   for predicting intended dexterous actions from videos; \emph{3)} we present a method for  estimating tactile signals from visual input without considering a model of the object; \emph{4)}  we provide new datasets that serve as test-beds for the aforementioned tasks.



\section{Related work}\label{sec:related}
We will focus our review on studies along the following concepts: the idea of prediction, including prediction of intention and future events (\emph{a}), prediction beyond appearance (\emph{b}), and  prediction of contact forces on hands (\emph{c}), work on hand actions (\emph{d}),  manipulation datasets (\emph{e}) and   action classification as a continuous process using various kinds of techniques and different kinds of inputs (\emph{f}).

{\bf  Prediction of Action Intention and Future Events: }
A small number of works in Computer Vision have aimed to predict intended action from visual input. For example, \cite{joo2014visual} use a ranking SVM to predict the persuasive motivation (or the intention) of the photographer who captured an image. \cite{pirsiavash2014inferring} seek to infer the motivation of the person in the image by mining knowledge stored in a large corpus using natural language processing techniques.
 \cite{yang2015grasp} propose that the grasp type, which is recognized in single images using CNNs, reveals the general  category of a person's intended action. In \citep{koppula2013anticipating}, a temporal Conditional Random Field model is used to infer anticipated human activities by taking into consideration object affordances. 
Other works attempt to predict events in the future. For example, 
 \cite{kitani2012activity} use concept detectors to predict future trajectories in a surveillance videos. \citep{Fouhey14b} learn from sequences of abstract images the relative motion of objects observed in single images. \cite{walker2014patch} employ visual mid-level elements to learn from videos how to predict  possible  object trajectories  in single images. More recently, \cite{vondrick2016} learn  using  CNN feature representations how to predict from one frame in the video the  actions and objects  in a future frame.
Our study also is about prediction of future events using neural networks. But while the above studies attempt to learn abstract concepts for reasoning in a passive setting, our goal is to perform online prediction of specific actions from  video of the recent past.
 

{\bf Physics Beyond Appearance:}
Many recent approaches in Robotics and Computer Vision aim to infer physical properties beyond appearance models from visual inputs. \cite{xie2013inferring} propose that implicit information, such as functional objects, can be inferred from video.
\citep{zhu2015understanding} takes a task-oriented viewpoint and  models objects using a simulation engine.  The general idea of associating images with forces has previously been used for object manipulation. The technique is called vision-based force measurement, and refers to the estimation of  forces  according to the observed deformations of an object \citep{Greminger2004}. Along this idea, recently \cite{Aviles2014} proposed a method using an RNN for the classification of forces due to tissue deformation in robotic assisted surgery.

{\bf Inference of Manipulation Forces:} The first work
in the Computer Vision literature to simulate   contact  forces during hand-object interactions is  \citep{pham2015towards}. Using as input RGB data, a model-based tracker estimates the poses of the hand and a known object, from which then the contact points and the motion trajectory are derived. Next, the minimal contact forces (nominal forces) explaining the kinematic observations are  computed from the Newton-Euler dynamics solving a conic optimization. Humans typically  apply more than the minimal forces. 
These additional forces are learned using a neural network on data collected from subjects, where  the force sensors  are attached to the object. Another approach on contact force simulation is  due to \citep{Rogez_2015_ICCV}. The authors segment the hand from RGBD data in single egocentric views and classify the pose  into 71 functional grasp categories  as proposed  in \citep{liu2014}.  Classified poses are matched to a library of graphically created hand poses, and theses poses are associated with force vectors normal to the meshes at contact points. Thus the forces on the observed hand are obtained  by finding the closest matching synthetic model. Both of these prior  approaches derive the forces using model based-approaches. The forces are computed   from the contact points, the shape of the hand, and  dynamic observations. Furthermore, both use RGBD data, while ours is an end-to-end learning approach  using as input only images. 

{\bf Dexterous Actions:}
The robotics community has been studying  perception and control problems of dexterous actions for decades \citep{shimoga1996robot}.  Some  works have studied grasping taxonomies \citep{cutkosky1989grasp,Feix2009}, how to recognize grasp types \citep{Rogez_2015_ICCV} and how to encode and represent human hand motion \citep{Romero2013}. \cite{pieropan2013functional} proposed a representation of  objects  in terms of their interaction with human hands. Real-time visual trackers \citep{Oikonomidis2011} were developed, facilitating computational research with hands.   Recently, several learning based systems were reported that infer contact points or how to grasp an object from its  appearance \citep{saxena2008robotic,lenz2014deep}.

{\bf Manipulation Datasets:} A number of object manipulation datasets  have been created, many of them recorded with wearable cameras providing egocentric views. For example, the Yale grasping dataset \citep{bullock2015yale} contains  wide-angle head-mounted camera videos recorded from four people during regular activities with images  tagged with  the hand grasp (of 33 classes). Similarly, the UT Grasp dataset \citep{cai2015scalable} contains head-mounted camera video of people grasping objects on a table, and was tagged with grasps (of 17 classes). The  GTEA set \citep{fathi2011learning} has egocentric videos of household activities with the objects annotated. Other datasets have egocentric RGB-D videos. The UCI-EGO \citep{rogez20143d} features object manipulation scenes with annotation of the 3D hand poses, and the GUN-71 \citep{Rogez_2015_ICCV} features subjects grasping objects, where care was taken to have the same amount of data for each of the 71 grasp types. Our datasets, in contrast, are taken from the third-person viewpoint. While having less variation in the visual setting than most of the above datasets, it focuses on the dynamic aspects of different actions, which  manipulate the same objects.


{\bf Action Recognition as an Online Process: }
Action recognition has been  extensively studied.
However, few of the proposed methods treat action recognition as a continuous (in the online sense) process; typically, action classification is performed on \emph{whole} action sequences \citep{Schuldt04,Ijina2014}.
Recent works  include building robust action models based on MoCap data \citep{wang2014learning} or using CNNs for large-scale video classification \citep{karpathy2014large,simonyan2014very}.
Most methods that take into account action dynamics usually operate under a stochastic process formulation, e.g., by using Hidden Markov Models \citep{lv2006recognition} or semi-Markov models \citep{shi2011human}. HMMs  can  model relations between consecutive image frames, but they cannot be applied to high-dimensional feature vectors.
In \citep{fanello2013keep} the authors  propose an online action recognition method by means of SVM classification of sparsely coded features on a sliding temporal window.
Most of the above methods assume only short-time dependencies between frames, make restrictive assumptions about the Markovian order of the underlying processs and/or rely on global optimization over the whole sequence. 

In recent work a few studies proposed approaches to recognition of partially  observed actions under the headings of  {\it early event detection} or {\it early action recognition}. \cite{ryoo2011human}
creates  a representation that encodes how   histograms of spatio-temporal features change over time. In a probabilistic model, the histograms are modeled with Gaussian distributions, and MAP estimation  over  all subsequences is used to recognize the ongoing activity. A second approach in the paper models the sequential structure in the changing histogram representation, and matches subsequences of the video using dynamic programming. Both  approaches were evaluated on  full body action sequences. In \citep{ryoo2013first} images are represented by spatio-temporal features and histograms of optical flow, and a hierarchical structure of video-subsegments is used to detect partial action sequences in first-person videos. \cite{ryoo2015robot} perform early recognition of activities in first person-videos by capturing  special sub-sequences characteristic for the onset of the main activity. \cite{Hoai-DelaTorre-IJCV14} propose a maximum-margin framework (a variant of SVM) to train visual detectors to recognize partial events. The classifier is trained with all the  video  sub-sequences of different length. To enforce the sequential nature of the events, additional  constraints on the score function of  the classifier are enforced, for example,  it has to increase as more frames are matched. The technique was demonstrated in multiple applications, including detection of facial expressions, hand gestures, and activities.

The main learning tools used here, the RNN and the Long Short Term Memory (LSTM) model, were recently popularized in language processing, and have been used for translating videos to language \citep{Venugopalan2014}, image description generation \citep{donahue2015long}, object recognition \citep{Visin2015}, and the estimation of object motion \citep{Fragkiadaki2015}. RNNs were also used for  action recognition 
\citep{ng2015beyond} to learn dynamic changes within the action. The aforementioned paper still performs whole video classification by using average pooling
and does not consider the use of RNNs for prediction. In  a very recent work, however,  \cite{Ma_2016_CVPR} train a LSTM using novel ranking losses for early activity detection.
Our contribution regarding action recognition is not that we introduce a new technique. We use an existing method (LSTM) and demonstrate it in an online prediction system. The  system keeps predicting, and considers the prediction reliable, when the predicted label converges (i.e. stays the same over a number of frames).
Furthermore, the subject of our study is novel. The  previous approaches consider the classical full body action problem. Here our emphasis is specifically on the hand motion, not considering other information such as the objects involved.

\section{Our Approach}

In this section, we first review the basics of Recurrent Neural Networks (RNNs) and the  Long Short Term Memory (LSTM) model.
Then we describe the specific algorithms for prediction of actions and forces used in our approach.

\subsection{Recurrent Neural Networks}

Recurrent Neural Networks have long been used for modeling temporal sequences. The recurrent connections are feedback loops in the unfolded network, and because of these connections  RNNs are suitable for  modeling time series with strong nonlinear dynamics and long time correlations.

Given a sequence $x=\{ x_1, x_2, \ldots, x_T \}$, a RNN computes a sequence of hidden states $h=\{ h_1, h_2, \ldots, h_T \}$ and outputs $y=\{ y_1, y_2, \ldots, y_T \}$ as follows:
\begin{align}
 & h_t = \mathcal{H}(W_{ih}x_t + W_{hh}h_{t-1} + b_h) \\
 & y_t = \mathcal{O}(W_{ho}h_t + b_o),
\end{align}
  where $W_{ih}, W_{hh}, W_{ho}$ denote  weight matrices, $b_{h}, b_{o}$ denote the biases, and  $\mathcal{H}(\cdot)$ and $\mathcal{O}(\cdot)$ are the activation functions of the hidden layer and the output layer, respectively. Typically, the activation functions are defined as  logistic sigmoid functions.

The traditional RNN is hard to train due to the so called \emph{vanishing gradient} problem, i.e. the weight updates computed via error backpropagation through time may become very small. 
The Long Short Term Memory model \citep{hochreiter1997long} has been proposed as a solution to overcome this problem. The LSTM architecture uses memory cells with gated access to store and output information, which alleviates the vanishing gradient problem in backpropagation over multiple time steps.

\begin{figure}[t]
  \centering
    \includegraphics[width=0.9\columnwidth]{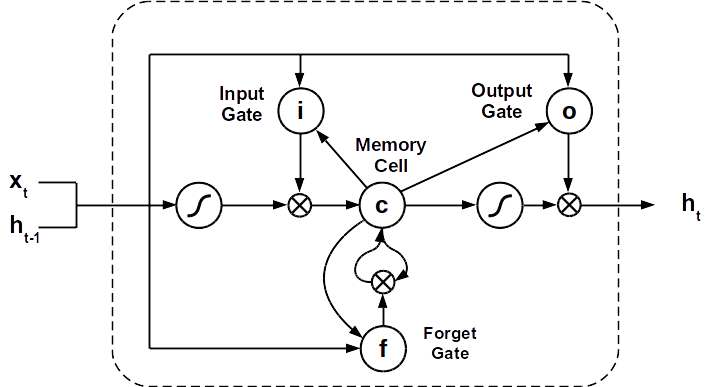}
  \caption{A diagram of a LSTM memory cell (adapted from \citep{graves2013speech}).}
  \label{fig:lstm_diagram}
\end{figure}

Specifically, in addition to the hidden state $h_t$, the LSTM also includes an input gate $i_t$, a forget gate $f_t$, an output gate $o_t$, and the memory cell $c_t$ (shown in Figure \ref{fig:lstm_diagram}).
The hidden layer and the additional gates and cells are updated as follows:
\begin{align}
 & i_t = \sigma(W_{xi}x_t + W_{hi}h_{t-1} + W_{ci}c_{t-1} + b_i)  \label{eq:lstm-first} \\
 & f_t = \sigma(W_{xf}x_t + W_{hf}h_{t-1} + W_{cf}c_{t-1} + b_f) \\
 & c_t = f_t c_{t-1} + i_t \tanh(W_{xc}x_t + W_{hc}h_{t-1} + b_c) \\
 & o_t = \sigma(W_{xo}x_t + W_{ho}h_{t-1} + W_{co}c_t + b_o) \\
 & h_t = o_t \tanh(c_t)
 \label{eq:lstm-last}
\end{align}

In this architecture, $i_t$ and $f_t$ are sigmoidal gating functions, and  these two terms learn to control the portions of the current input and the previous memory that the LSTM takes into consideration for overwriting the previous state. Meanwhile, the output gate $o_t$ controls how much of the memory should be transferred to the hidden state. These mechanisms allow LSTM networks to learn temporal dynamics with long time constants.

\subsection{RNN for action prediction}
\label{sec:action_prediction}
In this section, we describe our proposed model for action prediction.
We focus on manipulation actions where a person manipulates an object using a single hand.
Given a video sequence of a manipulation action, the goal is to generate a sequence of belief distributions over the predicted actions while watching the video.
Instead of assigning an action label to the whole sequence, we continuously update our prediction as frames of the video are processed.

\begin{figure}[th]
  \centering
    \includegraphics[width=0.8\columnwidth]{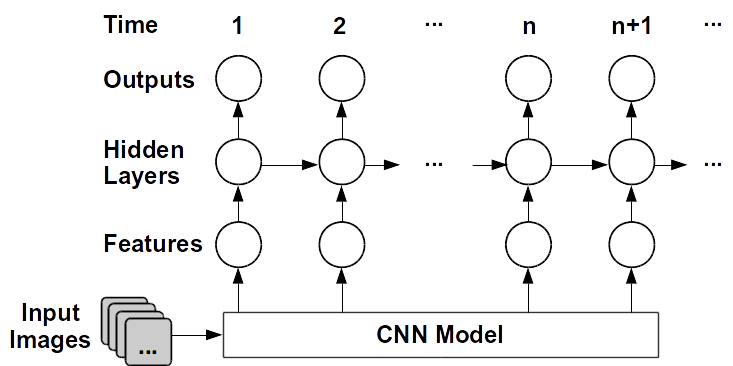}
  \caption{The flowchart of the action prediction model, where the LSTM model is unfolded over time.}
  \label{fig:lstm_unfolded}
\end{figure}

\paragraph{Visual representation:}
The visual information most   essential for manipulation actions comes from the pose and movement of the hands, while the body movements are less important. Therefore, we first track the hand using a mean-shift based tracker \citep{bradski1998computer}, and use cropped  image patches centered on  the hand.
In order to create abstract representations of image patches, we project each patch through a pre-trained CNN model (shown in Figure \ref{fig:lstm_unfolded}).
This provides the  feature vectors  used as input to the RNN.

\paragraph{Action prediction:}
In our model, the LSTM is trained using  as input a sequence of feature vectors $x=\{ x_1, x_2, \cdots, x_T \}$ and the action labels $y \in [1,N]$.
The hidden states and the memory cell values are updated according to equations (\ref{eq:lstm-first})-(\ref{eq:lstm-last}).
Then logistic regression is used to map the hidden states to the label space as follows:
\begin{equation}
P(Y=i|h_t, W_u, b_u) = softmax_{i}(W_{u}h_t + b_u).
\end{equation}
Then the predicted action label is obtained  as:
\begin{align}
\hat{y_t} = argmax_{i} P(Y=i|h_t, W_u, b_u).
\end{align}

\paragraph{Model learning:}
We follow the common approach of  training the model by minimizing the negative log-likelihood over the dataset $\mathcal{D}$.
The loss function is defined as 
\begin{align}
l(\mathcal{D}, W, b) = - \sum_{i=0}^{\vert \mathcal{D} \vert}
                                   \log (P(Y=y^{(i)}|x^{(i)}, W, b)),
\end{align}
where $W$ and $b$ denote the weight matrix and the bias term.
These parameters can be learnt using the stochastic gradient descent algorithm.

Since we  aim for the ongoing prediction rather than a classification of the whole sequence, we do not perform a pooling over the sequences to generate the outputs.
Each prediction is based  only on the current frame and the current hidden state, which implicitly encodes information about the history.
In practice, we achieve learning  by performing backpropagation at each frame. 

\subsection{RNN for prediction of forces at the fingers}
\label{sec:force_prediction}
We use a model similar to the one  above  to predict the forces  on the fingers from  visual input. Given  video sequences of actions, as well as  simultaneously recorded sequences of force measurements (see Sec.~\ref{sec:glove}), we reformulate the LSTM model, such that it predicts force estimates as close as possible to the ground truth values.

As before, we use as input to the LSTM features from pre-trained CNNs applied to image patches. In addition, the force measurements $v=\{ v_1, v_2, \cdots, v_T \}$, $v_t \in R^M$, are used as target values, where $M$ is the number of force sensors attached to the hand. Then the   forces are estimated as:
\begin{align}
\hat{v_t} = W_{v}h_t + b_v.
\end{align}

To train the force estimation model, we define the loss function as the least squares distance between the estimated value and the ground truth, and minimize it over the training set using stochastic gradient descent as:
\begin{align}
l(\mathcal{D}, W, b) = \sum_{i=0}^{\vert \mathcal{D} \vert} \sum_{t=0}^{T} {\Vert \hat{v_t} - v_t \Vert}_2^{2}
\end{align}

\section{Data collection} \label{sec:data_collection}

\subsection{A device for  capturing finger  forces during  manipulation actions}\label{sec:glove}


\begin{figure}[b]
  \centering
 \begin{tabular}{cc}
   \includegraphics[height=3.2cm]{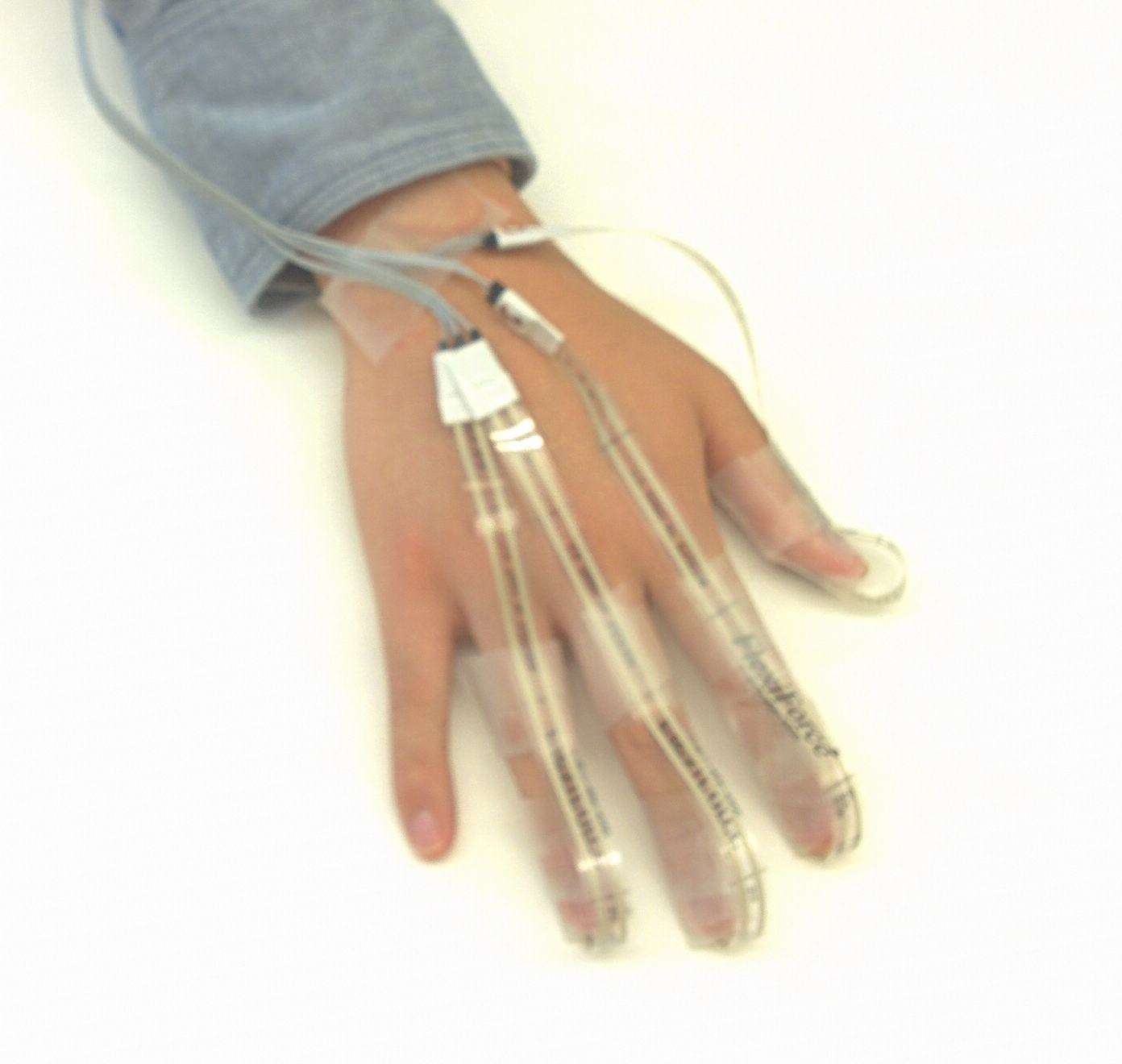} &
   \includegraphics[height=3.2cm]{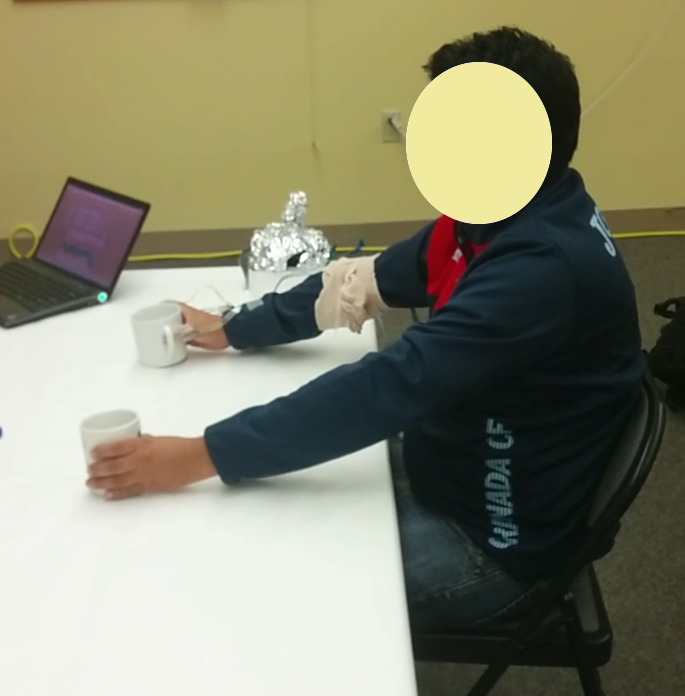} \\
   (a) & (b) 
 \end{tabular}
  \caption{Illustration of the force-sensing device. (a) The  sensors on four fingers; (b) The data collection. }
  \label{fig:gloves}
\end{figure}

We made a force sensing device with four force sensors attached directly to four fingers: the thumb, the pointer, the middle and the ring finger (see Figure \ref{fig:gloves}(a)). We omitted the  small finger, as the forces on this finger are usually quite small and not consistent across subjects (as found also by \citep{pham2015towards}).
We used the piezoresistive force sensors by Tekscan, with a documented accuracy (by the manufacturer) of $\pm 3\%$. The sensors at the finger tips have a measurement range of $0$ to $8.896$ N ($2$ lb), with a round sensing area of 9.53 mm in diameter. 
The entire sensing area is treated as one single contact point. 

\begin{figure}[th]
  \centering
 \begin{tabular}{cc}
   \includegraphics[width=0.45\columnwidth]{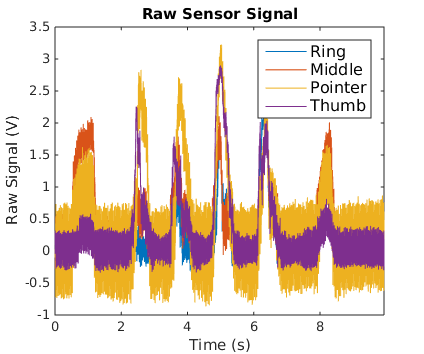} &
   \includegraphics[width=0.45\columnwidth]{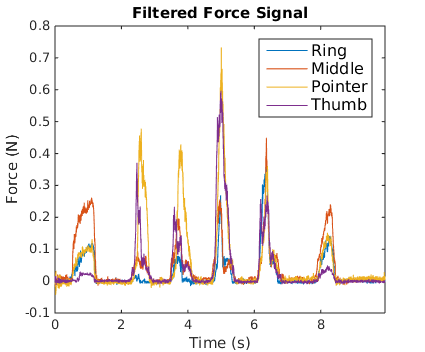} \\
   (a) & (b) 
 \end{tabular}
  \caption{Force data collection. (a) The raw, unfiltered voltage signal from the fingertip force sensors. (b) The filtered force signal from the fingertip sensors. }
  \label{fig:force_data_sample}
\end{figure}

The raw sensor outputs are voltages, from which  we derived the   forces perpendicular to the sensor surfaces as: 
\begin{equation}
F=4.448 * \left( C_1 * \frac{V_{out}}{V_{in} - V_{out}} - C_2 \right) ,
\end{equation}
where $V_{out}$ is the sensor measurement. $V_{in}$, $C_1$, and $C_2$ are fixed constants of the system. 
To remove environmental noise, we applied notch filtering  to the raw data, which gave us clear and smooth force outputs (see Figure \ref{fig:force_data_sample}).
The software, which  we designed for the device, will be released as a ROS package, including data recording and force visualization modules. 

\subsection{Manipulation Datasets}
Two datasets were collected. The first dataset
contains videos of  people performing dexterous   actions on various  objects. The  focus was to have different  actions (with significant variation) on the same object. This dataset was used to validate our approach of visual  action prediction. 

The second dataset
contains simultaneously recorded video and force data streams, but it has fewer objects. It was used to evaluate our approach of hand force estimation. 

\subsubsection{Manipulation action dataset (MAD)}

We asked five subjects to perform a number of actions with five objects, namely \textit{cup}, \textit{stone}, \textit{sponge}, \textit{spoon}, and \textit{knife}. Each object was manipulated in five different actions with five repetitions, resulting in a total of $625$ action samples. Table.~\ref{tbl:cls} lists all the object and action pairs considered in MAD. 

\begin{table}[ht]
\caption{Object and Action pairs of MAD}
\label{tbl:cls}
\begin{center}
\begin{tabular}{|l|l|}
\hline
 Object &  Actions \\ 
 \hline 
 cup & drink, pound,shake,move,pour  \\
 stone & pound,move,play,grind,carve \\
 sponge & squeeze,flip,wash,wipe,scratch \\
 spoon & scoop,stir,hit,eat,sprinkle \\
 knife & cut,chop,poke a hole,peel,spread \\ 
 \hline 

\end{tabular}
\end{center}
\end{table}

Since our aim was to build a system that can predict the  action as early as possible, we wanted to study the prediction performance during different phases in the action. 
To facilitate such studies, we  labeled the time in the videos when the hand establishes contact with the objects, which we call the ``touching point.''

\subsubsection{Hand actions with force dataset (HAF)}

To solve the problem of synchronization, we asked subjects to  wear on their right hand the force sensing device, leave  their  left hand bare, and  then perform with both hands the same action, with one hand mirroring the other (see Figure \ref{fig:gloves}(b) for the setting). 
We recorded from  five subjects performing different manipulation actions on four objects, as listed in Table \ref{tbl:haf_cls}. Each action was performed with five repetitions, resulting in a total of  $500$ sample sequences.

\begin{table}[ht]
\caption{Object and Action pairs of HAF}
\label{tbl:haf_cls}
\begin{center}
\begin{tabular}{|l|l|}
\hline
 Object &  Actions \\ 
 \hline
 cup & drink, move, pound, pour, shake  \\
 fork & eat, poke a hole, pick, scratch, whisk \\
 knife & chop, cut, poke a hole, scratch, spread \\ 
 sponge & flip, scratch, squeeze, wash, wipe \\
 \hline 

\end{tabular}
\end{center}
\end{table}
\vspace{-5mm}

\section{An experimental study with humans}

We were interested in how humans perform 
in prediction at different phases during the action. 
Intuitively, we would expect that the hand configuration and motion just before the grasping of the object, when establishing contact, and shortly after the contact point can be very informative of the intended action. 
Therefore, in order to evaluate how early we can accurately predict, we investigated the prediction performance at certain time offsets with respect to the touching point.

\begin{figure}[ht]
  \centering
    \includegraphics[width=0.8\columnwidth]{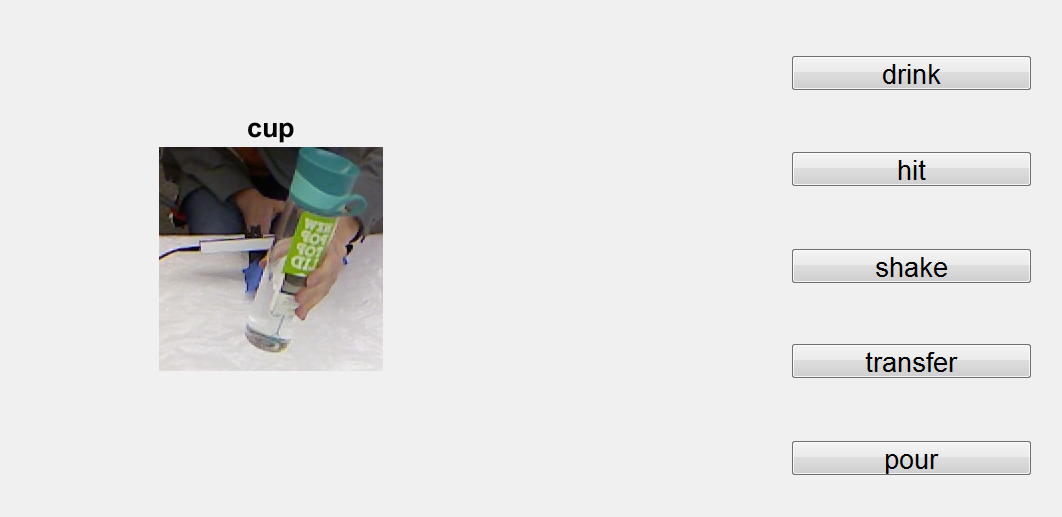}
  \caption{Interface used in the human study.}
  \label{fig:gui}
\end{figure}

We picked three objects from the MAD dataset for the study, namely \emph{cup}, \emph{sponge} and \emph{spoon}. The prediction accuracy at four different time points was then evaluated: $10$ frames before the contact point, exactly at contact, $10$, and $25$ frames after the contact point.  Figure \ref{fig:gui} shows the interface subjects used in this study.

\begin{figure}[htbp]
  \centering
  \begin{tabular}{ll}
  \begin{minipage}{0.46\columnwidth}
  \begin{tabular}{c}
    \includegraphics[width=0.98\columnwidth]{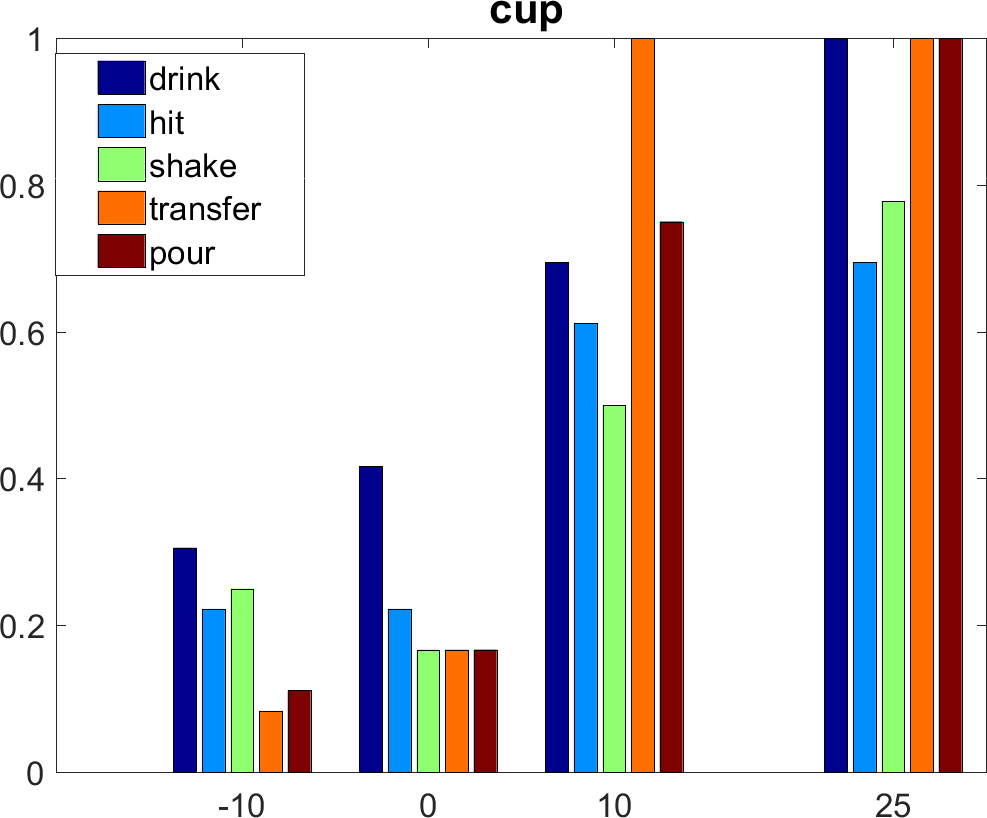}\\
    \includegraphics[width=0.98\columnwidth]{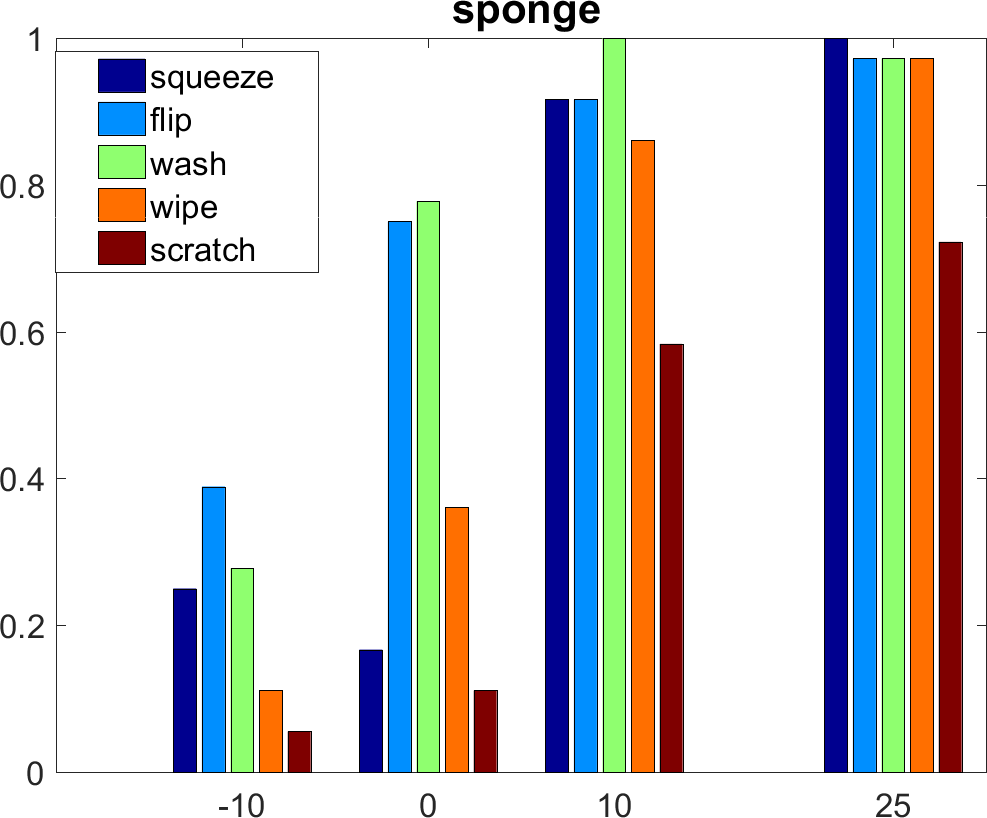} \\
    \includegraphics[width=0.98\columnwidth]{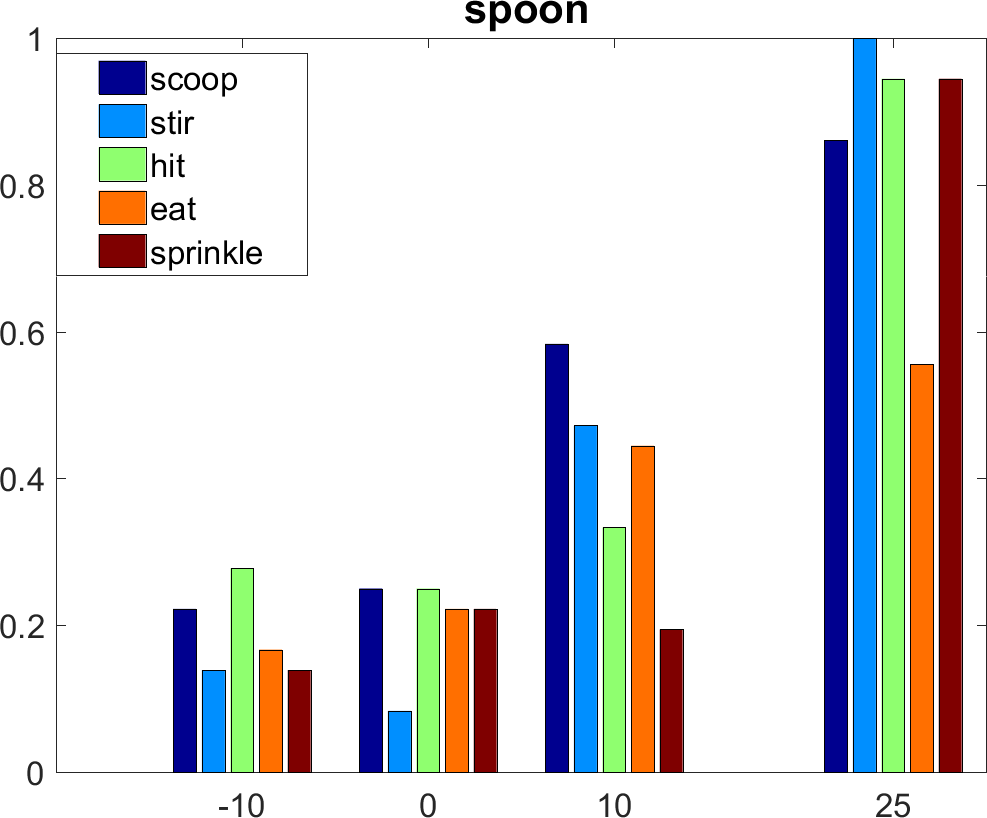} \\
    (a) without feedback
    \end{tabular}
    \end{minipage} &
  \begin{minipage}{0.46\columnwidth}
  \begin{tabular}{c}
    \includegraphics[width=0.98\columnwidth]{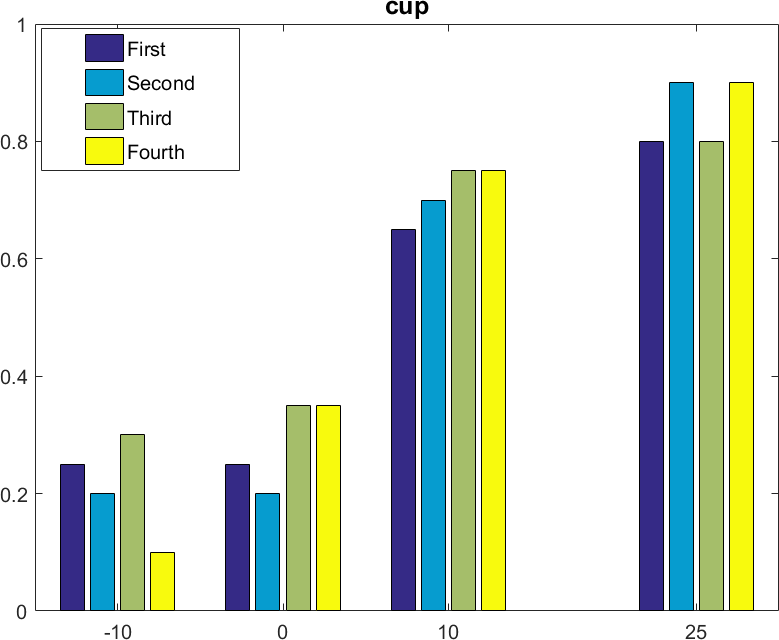} \\
    \includegraphics[width=0.98\columnwidth]{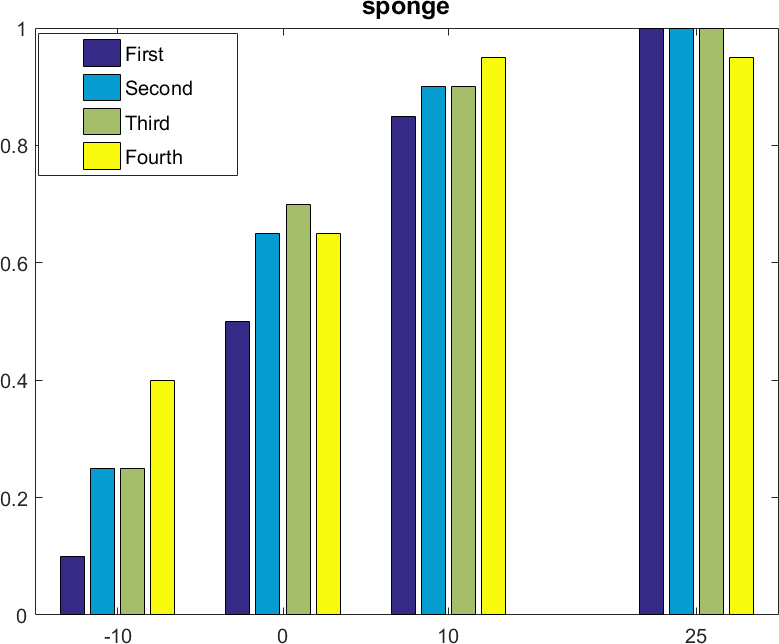}\\   
    \includegraphics[width=0.98\columnwidth]{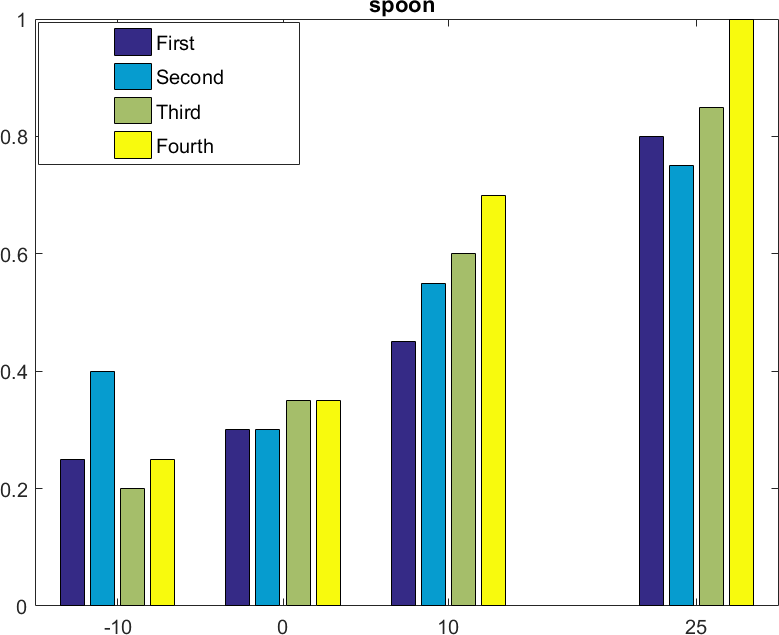}\\
    (b) with feedback
    \end{tabular}
    \end{minipage}
    \end{tabular}
  \caption{Human prediction performance. (a) First study (without feedback). Success rate for three objects (cup, sponge, and spoon) for  five different actions at four time offsets. (b) Second study (with feedback). Success rate for three objects averaged over five actions over four sets of videos at four offsets.}
  \label{fig:hum_exp}
\end{figure}

In a first experiment we asked 18 human subjects to perform the prediction task. For each of the three objects, after a short ``training'' phase in which all actions were demonstrated at full length, each subject was shown a set of 40 video segments and was asked to identify the currently perceived action. Each segment ended at one of the four time points relative to the contact point described above and was constructed from the same hand patches used in the computational experiments. All actions and all time offsets were equally represented. Figure \ref{fig:hum_exp}(a) plots subjects' average prediction performance  for the different  objects, actions and  time offsets. With
five actions per object, 20\% accuracy corresponds to chance level. As we can see, the task of judging before and even at contact point, was very difficult and  classification  was at chance for two of
the objects, the spoon and and the cup, and above chance
at contact only for the sponge. At 10 frames after contact human classification becomes better and reaches in average about 75\% for the sponge, 60\% for the cup, but only 40\% for the spoon. At 25 frames subjects' judgment becomes quite good with the sponge going above  95\% for four of the five actions, and the other two actions  in average at about 85\%. We can also see which actions are easily confused. For the cup, 'shake' and 'hit' were  even after 25 frames still difficult to recognize, and for the spoon the early phases of movement for most actions appeared  similar, and 'eat' was most difficult to identify.

To see whether there is additional distinctive information in the actors' movement, and subjects can take advantage of it with further learning, we performed a second study. Five participating subjects  were shown 4 sets of  40 videos for each object, and this time they were given feedback on which was the correct action. Figure \ref{fig:hum_exp}(b) shows the overall success rate for each object and time offset over the four sets. If learning occurs, subjects' should improve from the first  to the fourth set. The graphs show that there is a  bit of learning. The effect is largest for  the spoon, where subjects can learn to better distinguish at 10 frames after contact.
The  focus was to have different  actions (with significant variation) on the same object.

\begin{figure*}
  \centering
 \includegraphics[width=0.95\textwidth]{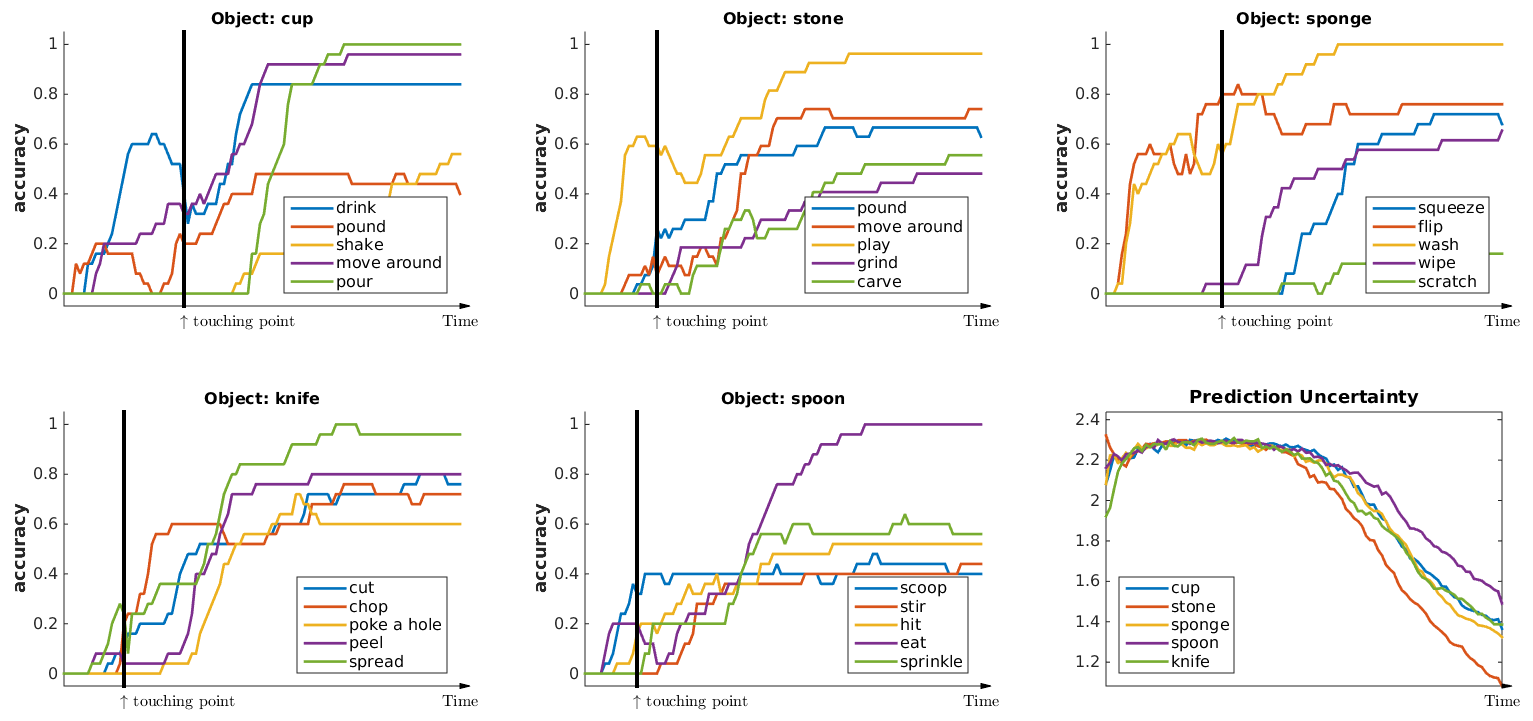}
  \caption{Prediction accuracies over time for the five different objects, and Prediction Uncertainty computed from the entropy. The black vertical bars show the touching point. For each object we warped and aligned all the sample sequences so that they align at  the same touching point. Best viewed in color.}
  \label{fig:pred_accu}
\end{figure*}

\section{Experimental results}

The two algorithms have been implemented in a system that runs in  real-time on a GPU.  This sections reports
three experimental evaluations.
The first experiment evaluates the prediction performance as an on-going task, the second  compares our action recognition algorithm against human performance, and  the third  evaluates our force estimation. 

\subsection{Hand action prediction on MAD}


Our approach uses visual features obtained with deep learning, which serve as input to a sequence learning technique. 

First we apply the mean-shift based tracker of \cite{comaniciu2000real}  to obtain the locations of the hand. We crop image patches of size  $224 \times 224$ pixels, centered on the hand. Then
our feature vectors are  computed by projecting these patches  through a convolutional neural network. To be specific, we employ the VGG network \citep{Simonyan14c} with 16 layers, which has been pre-trained on the ImageNet. We take the output of layer  ``fc7''  as feature vector ($4096$ dimensions), which we then use to train a one layer LSTM RNN model for action prediction.

Our RNN has hidden states of $64$ dimensions, with all the weight matrices randomly intialized  using the normal distribution. 
We first learn a linear projection to map the $4096$ input features to the  $64$ dimensions of the RNN.
We use mini-batches of $10$ samples and the adaptive learning rate method to update the parameters. The training stops after $100$ epochs in  all the experiments.

To evaluate the action prediction performance, we performed leave-one-subject-out cross-validation over the five subjects.
Each time we used the data from one subject for testing, and trained the model on the  other four subjects. Then all the results were averaged over the five rounds of testing.

\subsubsection{On-going prediction}
\label{sec:ongoing}

Our goal is to understand how the recognition of action improves over time. 
Thus, we plot the prediction accuracy as a function of time, from  the action preparation to the end of the action. Our system performs predictions based on every new incoming frame as the action unfolds. 





The first five plots in Figure \ref{fig:pred_accu} show the change in  prediction accuracy over time. For a given action video, our system generates for each frame a potential score vector (with one value for each action)  to form a score sequence of  same length as the input video. 
Since the actions have different length, we  aligned them at the touching points. To be specific, we resampled the sequences before and after the touching points to the same length. For each object, we show the prediction accuracy curves of the five actions. 

The vertical bar in each figure indicates the time of the \textit{touching point}. The touching point splits the sequence into two phases: the ``preparation'' and the ``execution''. It is interesting to see that for some object-action  pairs our system yields  high prediction accuracy even before the touching point, e.g. the ``cup - drink'' and ``sponge - wash''.

The last plot in Figure \ref{fig:pred_accu}
shows the change of prediction uncertainty over time for each of the five objects. This measure was derived  from the entropy over the different actions. As can be seen, in all cases, the uncertainty drops rapidly as the prediction accuracy rises along time. 

\subsubsection{Classification results}
At the end of the action, the on-going prediction task becomes a  traditional classification. To allow evaluating  our method on  classical action recognition, we also computed  the classification results for the whole video. The estimate over the sequence was derived as a weighted average over all frames using a linear weighting with largest value  at the last frame. To be consistent with the above, the classification was performed for each object  over the five actions considered.

\begin{figure}[ht]
  \centering
    \includegraphics[width=\columnwidth]{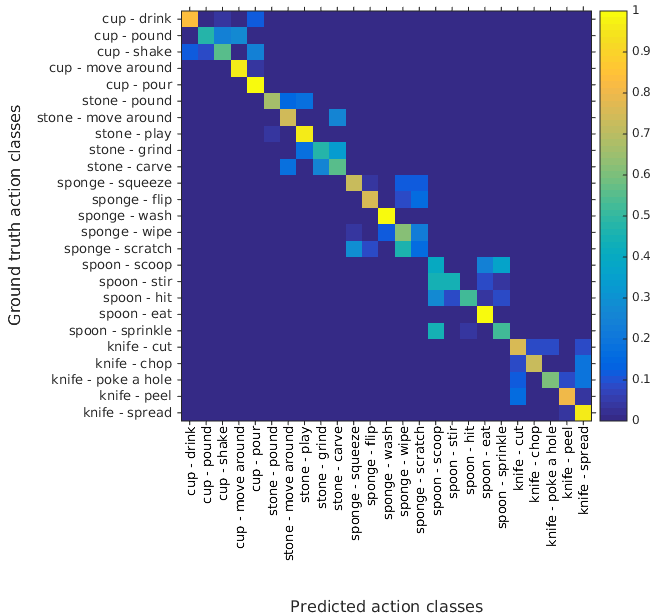}
  \caption{Confusion matrix of action classification.}
  \label{fig:action_cm}
  \vspace{-20pt}
\end{figure}

Figure \ref{fig:action_cm} shows the confusion matrix of the action classification results.
One can see that, our model achieved high accuracy on various object-action combinations, such as ``cup /drink'' and ``sponge/wash'', where the precision exceeds $90 \%$.

We used two traditional classification methods as our baseline: Support Vector Machine (SVM) and Hidden Markov Model (HMM). For the HMM model, we used the mixture of Gaussian assumption and we chose the number of hidden states as five. Since the SVM model doesn't accept input samples of different length, we used a sliding window ($size=36$) mechanism. We performed  the classification over each window, and then combined the results using majority voting. For both these baseline methods, we conducted a dimension reduction step to map the input feature vectors to 128 dimensions using PCA. To further explore the efficiency of the LSTM method in  predicting actions on our dataset, we also applied the LSTM model using HoG features as input. The average accuracy was found $59.2\%$, which is $10\%$ higher than the HMM and $23\%$ higher than the SVM, but still significantly lower than our proposed method.

\begin{table}[ht]
\caption{Comparison of classification accuracies on different objects}
\label{tbl:cls-accu}
\begin{center}
\begin{tabular}{|l|cccc|}
\hline

 \multirow{2}{*}{Object/Action} & \multirow{2}{*}{SVM} & \multirow{2}{*}{HMM} & LSTM & LSTM \\
   &  &  & HOG & VGG16 \\
 \hline 
 cup/drink &  79.1\% &  96.0\% &  82.9\% &  92.5\% \\ 
 cup/pound &  20.0\% &  81.7\% &  40.0\% &  73.3\% \\ 
 cup/shake &  64.3\% &  56.8\% &  32.6\% &  83.3\% \\ 
 cup/move &  62.7\% &  53.2\% &  51.9\% &  82.1\% \\ 
 cup/pour &  60.0\% &  100.0\% &  80.3\% &  80.8\% \\ 
 \hline
 stone/pound &  26.7\% &  73.3\% &  60.0\% &  73.3\% \\ 
 stone/move &  87.8\% &  68.0\% &  90.0\% &  61.4\% \\ 
 stone/play &  64.6\% &  97.1\% &  60.5\% &  86.7\% \\ 
 stone/grind &  28.3\% &  45.0\% &  60.0\% &  46.7\% \\ 
 stone/carve &  43.3\% &  28.5\% &  66.0\% &  39.1\% \\ 
 \hline
 sponge/squeeze &  41.1\% &  81.7\% &  64.3\% &  83.4\% \\ 
 sponge/flip &  53.3\% &  91.0\% &  96.0\% &  71.0\% \\ 
 sponge/wash &  85.9\% &  84.6\% &  91.1\% &  92.5\% \\ 
 sponge/wipe &  46.9\% &  47.5\% &  58.1\% &  46.3\% \\ 
 sponge/scratch &  30.0\% &  0.0\% &  43.3\% &  15.0\% \\ 
 \hline
 spoon/scoop &  39.0\% &  27.1\% &  53.6\% &  32.0\% \\ 
 spoon/stir &  45.3\% &  30.0\% &  20.0\% &  74.3\% \\ 
 spoon/hit &  28.9\% &  20.0\% &  22.4\% &  56.7\% \\ 
 spoon/eat &  65.0\% &  79.2\% &  78.1\% &  81.1\% \\ 
 spoon/sprinkle &  60.0\% &  25.0\% &  40.5\% &  69.1\% \\ 
 \hline
 knife/cut &  33.5\% &  33.7\% &  49.6\% &  75.3\% \\ 
 knife/chop &  0.0\% &  45.0\% &  43.3\% &  72.7\% \\ 
 knife/poke a hole &  33.3\% &  20.0\% &  51.0\% &  72.0\% \\ 
 knife/peel &  66.3\% &  28.9\% &  90.0\% &  72.5\% \\ 
 knife/spread &  38.2\% &  28.3\% &  54.3\% &  74.2\% \\ 
 \hline 
 Avg. &  48.1\%  &  53.7\%  &  59.2\%  &  68.3\% \\ 
 
\hline
\end{tabular}
\end{center}
\end{table}

\subsubsection{Discussion}
It should be noted that this is a novel, challenging dataset with no equivalent publicly available counterparts. Subjects performed the action in unconstrained conditions, and thus there was a lot of variation in their movement, and they performed some of the actions in very similar ways, making them difficult to distinguish, as  also  our human study confirms.

The results demonstrate that deep learning based continuous  recognition of  manipulation actions is feasible, providing a promising alternative to traditional methods such as HMM, SVM and other methods  based on hand-crafted features.

\subsection{Action prediction at the point of contact, before and after}

\begin{figure}[th]
  \centering
  \includegraphics[width=\columnwidth]{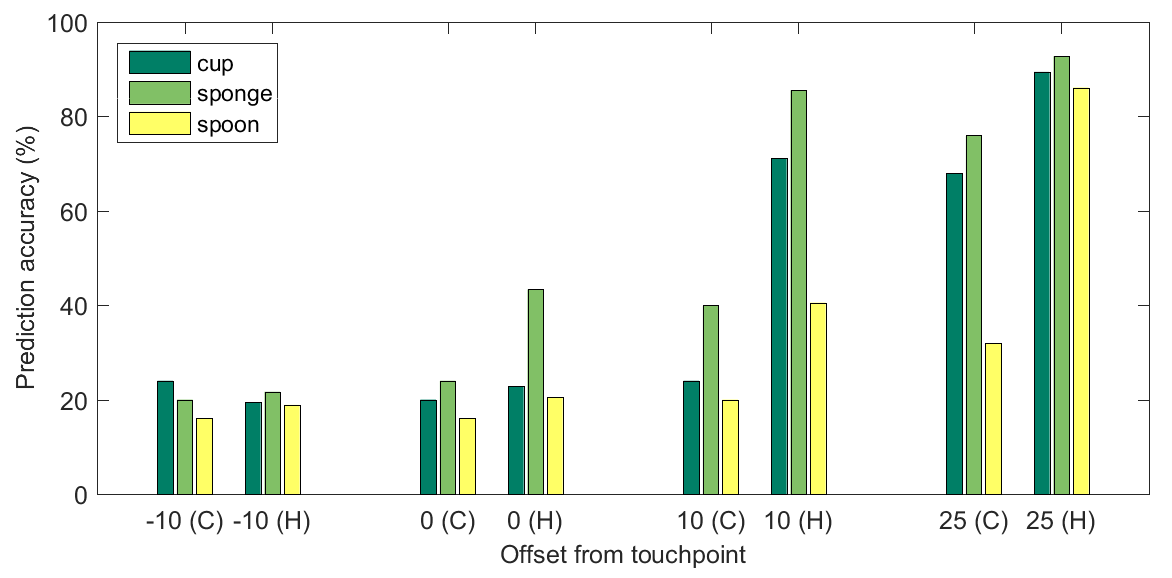}
  \caption{Comparison of prediction accuracies between our computational method ({\bf C}) and data from  human observers ({\bf H}). Actions are classified at four different time points before, at, and after the touching point (at -10,0,+10, +25 frames from the touching point). C denotes the learnt model, H denotes the psychophysical data).}
  \label{fig:touching-point-analysis}
\end{figure}

We next compare the performance of  our online algorithm (as evaluated in Section \ref{sec:ongoing}) against those of human subjects.
Figure \ref{fig:touching-point-analysis} summarizes the prediction performance per object and time offset.
As we can see our algorithm's performance is not significantly behind those of humans.
At ten frames after contact, computer  lags behind human performance. However, at 25 frames after the contact point, the gaps between our proposed model and human subjects are fairly small. 
Our model performs worse on the spoon, but this is likely due to the large variation in the way different people move  this object. Our human study already revealed the difficulty in judging spoon actions, but the videos shown to subjects featured less actors than were tested with the algorithm.
Considering this, we can conclude that our algorithm is already close to human performance in  fast action prediction. 

\subsection{Hand force estimation on HAF}

\begin{figure}[ht]
\centering
 \includegraphics[width=1\columnwidth]{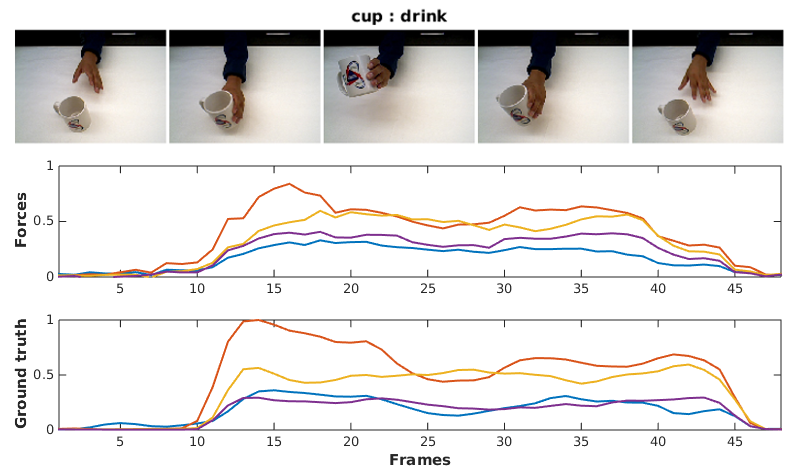}
 \vspace{-0.3cm}
  \caption{Illustration of the online force estimation system. The video frames in the top row show samples of the action 'drinking from a cup.' 
The second row shows the estimated forces and the third row  the corresponding ground truth.}
  \label{fig:force_estimation_illus}
\end{figure}

In the following we demonstrate the ability of the RNN to predict the forces on the fingers directly from images.
We developed an online force estimation system. While watching a person performing  actions in front of the camera, our system provides the finger forces in real time. Figure \ref{fig:force_estimation_illus} shows one example of online force prediction for the ``drink'' action.
We next describe our training method, and then present our results.

\subsubsection{Training}
The  LSTM model ( described in Section \ref{sec:force_prediction}) is used to estimate the hand forces for each frame.
Since people have different  preferences in performing actions, the absolute force values can vary  significantly for the same action. 
Therefore, we first normalize the force samples, which are used for training, to the range $[0, 1]$.
The visual features in  the  video frames are obtained  the same way  as in the action prediction. 
Our LSTM model has one layer with 128 hidden states. To effectively train the model, we use the adaptive  learning rate method for updating the neural network, and we use a batch size of 10 and  stop the training at 100 epochs.

\subsubsection{Results}
\begin{figure*}[htbp]
\centering
 \includegraphics[width=1.9\columnwidth]{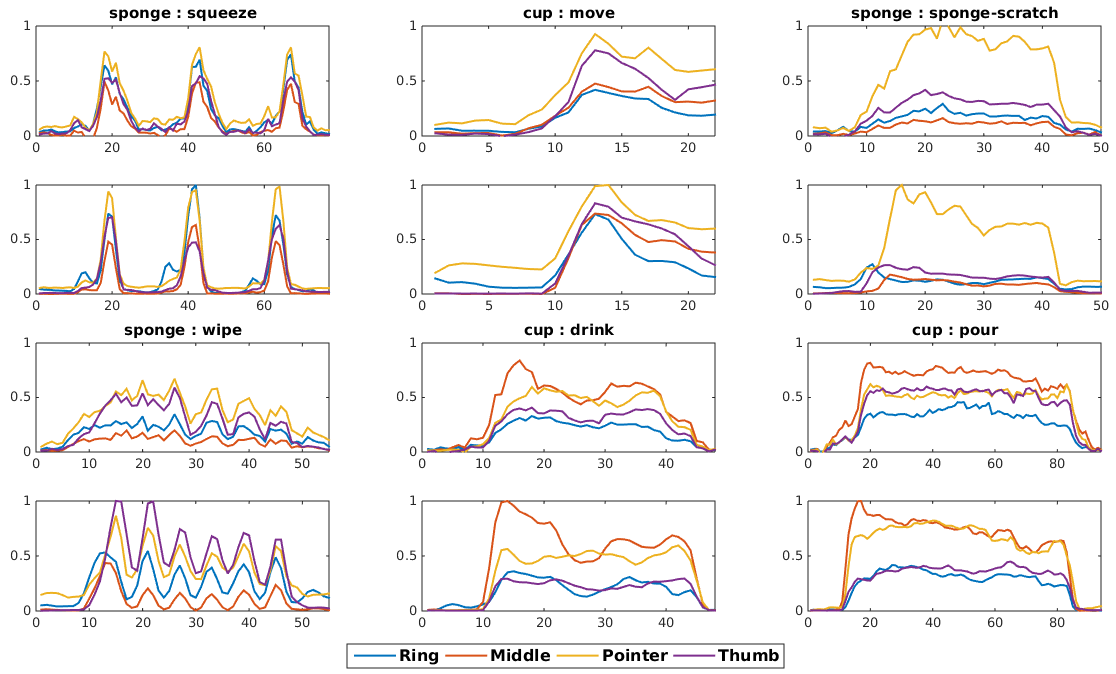}
  \caption{Samples of force estimation results. The first and third row show the  force estimation. The second and fourth row show the corresponding ground truth values.}
  \label{fig:force_comp}
\end{figure*}

We first show examples of our force estimation and then report the average errors. 
Figure \ref{fig:force_comp} shows sample results. For each of the six pairs, the upper graph shows the estimated forces, and  the lower one shows the ground truth. It can be seen that our system estimates well the overall force patterns for different actions. For example, for the ``sponge/squeeze'' action, the estimated forces correctly reproduce the three peaks of the real action, or for the ``cup/move'' action, the output forces predict the  much smoother changes. 
Table \ref{tbl:avg-err-finger} provides the average error of estimated force for each finger, and  Table \ref{tbl:avg-err-action} gives  the average estimation error for all the  actions. The errors are in the range of 0.075 to 0.155, which demonstrates that the method also has good quantitative prediction and potential for visual force prediction.

\begin{table}[ht]
\caption{Average errors of estimated force for each finger}
\label{tbl:avg-err-finger}
\begin{center}
\begin{tabular}{|l|cccc|}
\hline

   & Ring & Middle & Pointer  &  Thumb \\ 
\hline 
Avg.   &  0.103  &  0.098  &  0.130  &  0.119  \\  

\hline
\end{tabular}
\end{center}
\end{table}

\begin{table}[ht]
\caption{Average errors of estimated force for each action}
\label{tbl:avg-err-action}
\begin{center}
\begin{tabular}{|l|ccccc|}
\hline

 Object  & Action 1 & Action 2 & Action 3 & Action 4 & Action 5 \\ 
\hline
Cup   & Drink & Move & Pound & Pour & Shake \\
\hline
      & 0.096  & 0.122  & 0.108  & 0.107  & 0.110 \\
\hline
Fork   & Eat & Hole & Pick & Scratch & Whisk \\
\hline
      & 0.106  & 0.090  & 0.075  & 0.094  & 0.100 \\
\hline
Knife &  Chop & Cut & Poke & Scratch & Spread \\
\hline
      & 0.157  & 0.155  & 0.109 &  0.123  & 0.110 \\
\hline
Sponge & Flip & Scratch & Squeeze & Wash & Wipe \\
\hline
      & 0.101  & 0.130  & 0.112  & 0.127  & 0.121 \\
\hline

\end{tabular}
\end{center}
\end{table}

\begin{table}[ht]
\caption{Action prediction accuracy. Comparsion of prediction using vision data only ("Vision") against  using  vision and force data ("V+F").}
\label{tbl:pred_with_force}
\begin{center}
\begin{tabular}{|l|c|c|c|c|c|c|}
\hline

Object & cup  & stone  & sponge  & spoon  & knife  & Avg. \\ 
\hline
 Vision & 82.4\%  & 61.4\%  & 61.6\%  & 62.6\%  & 73.3\%  & 68.3\% \\ 
 \hline 
 V + F & 88.2\%  & 75.1\%  & 59.1\%  & 57.5\%  & 72.7\%  & 70.5\% \\ 
 \hline 
 
\end{tabular}
\end{center}
\end{table}

\subsubsection{Why predict forces?}
\label{sec:predict_forces}
One motivation for predicting forces, is that the additional data, which we learned through association, may help increase recognition accuracy. There is evidence that human understand others' actions in terms of their own motor primitives \citep{Gallese98,rizzolatti2001neurophysiological}. However, so far these findings have not been  modeled in  computational terms.  

To evaluate the usefulness of the predicted forces, we applied our force estimation algorithm on the MAF dataset to compute the force values. Then we used the vision data together with the regressed force values as bimodal information to train a network for  action predicton. Table \ref{tbl:pred_with_force} shows the results of the prediction accuracy with the bimodal information on different objects. Referring to the table, the overall average accuracy for the combined vision force data (V+F) was $2.2\%$ higher than  for  vision information only. This first attempt on predicting with bimodal data demonstrates the potential of utilizing visually estimated forces for recognition. Future work will further elaborate on the idea and explore  networks \citep{Hoffman_CVPR2016}, which can be trained  from both vision and force at the same time to learn "hallucinate" the forces and predict actions. 


As  discussed in the introduction, the other advantage is that  we will be able to teach robots through video demonstration. 
If we can predict forces exerted by the human demonstrator and provide the force profile of the task using  vision only, this would have a huge impact on the way robots learn  force interaction tasks.  
In  future work we plan to develop and employ sensors that can also measure the tangential forces, i.e. the frictions, on the fingers. We  also will expand the sensor coverage to the whole hand. With these two improvements, our method could be applied to a range of complicated task such as screwing or assembling. 


\section{Conclusion and Future work}
In this paper we proposed an approach to action interpretation, which treats the problem as a continuous updating of beliefs and predictions.
The ideas were  implemented for two tasks: the prediction of perceived action from visual input, and the prediction of force values on the hand. The methods were shown to run in real-time and demonstrated high accuracy performance. The action prediction was evaluated also against human performance, and shown to be nearly on par. Additionally, new datasets of videos of dexterous actions and force measurements were created, which can be accessed from \citep{Website}. 

The methods presented here are only a first implementation of a concept that can be further developed along a number of directions. Here, we applied learning on 2D images only, and clearly, this way we also learn properties of the images that are not relevant to the task, such as the background textures. In order to become robust to these `nuisances', 3D information, such as contours and depth features, could be considered in future work.
While the current implementation only considers action labels, the same framework can be applied for other aspects of action understanding. For example,  one can describe the different phases of actions and predict these sub-actions  since different actions share similar small movements. One can also describe the movements of other body parts, e.g., the arms and shoulders. Finally, the predicted forces may be used for learning how to perform actions on the robot. Future work will attempt to map the forces from the human hands onto other actuators, for example three-fingered hands or grippers.




 \section*{Acknowledgement}
 This work was funded by the support of the National Science
Foundation under grant SMA 1540917 and grant CNS
1544797, by Samsung under the GRO program (N020477,
355022), and by DARPA
through U.S. Army grant W911NF-14-1-0384.


\small{
\bibliographystyle{spbasic}      
\vspace{-0.5cm}
\bibliography{force_estimation}
}

%
%

\end{document}